\documentclass[lettersize,journal]{IEEEtran}
\usepackage{amsmath,amsfonts}
\usepackage{algorithmic}
\usepackage{algorithm}
\usepackage{array}
\usepackage[caption=false,font=normalsize,labelfont=sf,textfont=sf]{subfig}
\usepackage{textcomp}
\usepackage{stfloats}
\usepackage{url}
\usepackage{verbatim}
\usepackage{graphicx}
\usepackage{cite}
\usepackage{multirow}
\usepackage{float}
\usepackage{color}
\usepackage{graphicx}
\usepackage{amsmath}
\usepackage{amsthm}
\usepackage{booktabs}
\usepackage{bbding}
\usepackage{tikz,xcolor}

\usepackage[implicit=false]{hyperref}
\hypersetup{hidelinks,
	colorlinks=true,
	allcolors=black,
	pdfstartview=Fit,
	breaklinks=true}
\newcommand{\chy}[1]{\textcolor{blue}{#1}}

\definecolor{lime}{HTML}{A6CE39}
\DeclareRobustCommand{\orcidicon}{
\begin{tikzpicture}
\draw[lime, fill=lime] (0,0)
circle[radius=0.16]
node[white]{{\fontfamily{qag}\selectfont \tiny \.{I}D}};
\end{tikzpicture}
\hspace{-2mm}
}
\foreach \x in {A, ..., Z}{%
\expandafter\xdef\csname orcid\x\endcsname{\noexpand\href{https://orcid.org/\csname orcidauthor\x\endcsname}{\noexpand\orcidicon}}
}
\begin{document}

\title{Dynamic Textual Prompt For Rehearsal-free Lifelong Person Re-identification}

\author{%
Hongyu Chen, 
Bingliang Jiao\hspace{-1.5mm}\orcidC{}, %
Wenxuan Wang\orcidD{}, %
Peng Wang\hspace{-1.5mm}\orcidA{}%
\thanks{Hongyu Chen, Bingliang Jiao, Wenxuan Wang, Peng Wang, and Yanning Zhang are with the National Engineering Laboratory for Integrated Aero-Space-Ground-Ocean Big Data Application Technology, School of Computer Science, Northwestern Polytechnical University, Xi’an 710060, China. (e-mail: ChenHY@mail.nwpu.edu.cn; bingliang.jiao@mail.nwpu.edu.cn; wxwang.iris@gmail.com; peng.wang@nwpu.edu.cn; ynzhang@nwpu.edu.cn)
}
}



\maketitle

\begin{abstract} 
Lifelong person re-identification attempts to recognize people across cameras and integrate new knowledge from continuous data streams. Key challenges involve addressing catastrophic forgetting caused by parameter updating and domain shift, and maintaining performance in seen and unseen domains.
Many previous works rely on data memories to retain prior samples. However, the amount of retained data increases linearly with the number of training domains, leading to continually increasing memory consumption. 
Additionally, these methods may suffer significant performance degradation when data preservation is prohibited due to privacy concerns.
To address these limitations, we propose using textual descriptions as guidance to encourage the ReID model to learn cross-domain invariant features without retaining samples.
The key insight is that natural language can describe pedestrian instances with an invariant style, suggesting a shared textual space for any pedestrian images.  By leveraging this shared textual space as an anchor, we can prompt the ReID model to embed images from various domains into a unified semantic space, thereby alleviating catastrophic forgetting caused by domain shifts.
To achieve this, we introduce a task-driven dynamic textual prompt framework in this paper. 
This model features a dynamic prompt fusion module, which adaptively constructs and fuses two different textual prompts as anchors. This effectively guides the ReID model to embed images into a unified semantic space. Additionally, we design a text-visual feature alignment module to learn a more precise mapping between fine-grained visual and textual features.
We also developed a learnable knowledge distillation module that allows our model to dynamically balance retaining existing knowledge with acquiring new knowledge.
Extensive experiments demonstrate that our method remarkably outperforms SOTAs under various settings. Our model even outperforms rehearsal-based state-of-the-art approaches by $15.9\%/14.8\%$ over mAP/rank1 across several seen datasets and $17.4\%/17.9\%$ over mAP/rank1 on multiple unseen datasets over four training orders.

\end{abstract}

\begin{IEEEkeywords}
Person Re-Identification, Lifelong Learning, Dynamic Networks.
\end{IEEEkeywords}
\section{Introduction}

\IEEEPARstart{P}{erson} re-identification (ReID) endeavors to identify individuals across various cameras and viewpoints. Recently, this task has attracted more and more attention due to its potential applications in fields like social security, attributed to extensive datasets and continuous technique innovation. Nevertheless, practical ReID models must continually adapt to new datasets in real-world scenarios, necessitating a continuous learning approach to accommodate diverse data distributions.
Therefore, researchers have proposed the Lifelong Person Re-identification (LReID) task. 
The primary focus here is to mitigate catastrophic forgetting, where the model risks losing previously learned knowledge from seen domains when exposed to new datasets.
LReID assumes that only new datasets at the current moment are visible, while the previously trained seen datasets are no longer accessible for training.

\begin{figure}
  \centering
  \includegraphics[width=1.0\linewidth]{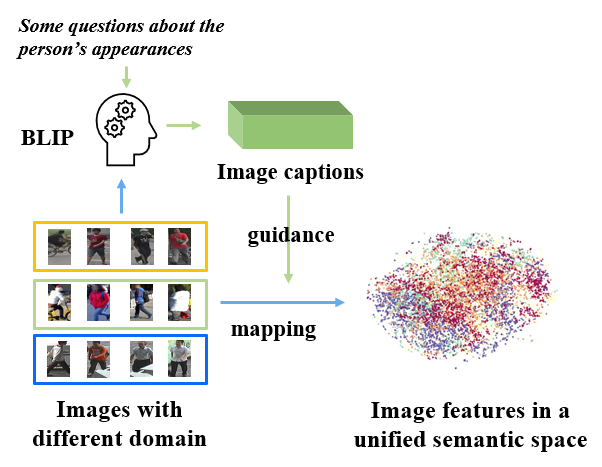}
   \caption{The inspirations of our rehearsal-free method. Natural language descriptions of pedestrian images from various domains act as domain-independent anchors. These consistent descriptions guide the mapping of images into a unified semantic space, effectively mitigating the issue of catastrophic forgetting.}
   \label{fig:intro}
\end{figure}

The challenges associated with the LReID ~\cite{AKA,MEGE} can be summarized as follows:
1) The model tends to adapt to the domain distribution of new incoming data, leading to forgetting the knowledge acquired from the previous datasets, \textit{i.e.}, catastrophic forgetting problem. 
2) Existing traditional lifelong learning tasks mainly focus on classification and emphasize inter-class differences. However, LReID requires capturing fine-grained discriminative features due to the significant intra-class similarities.
3) In practical scenarios, not every pedestrian has been observed previously. Therefore, the LReID model needs to maintain high performance across both seen and unseen domains.


Previous LReID methods predominantly rely on rehearsal strategy~\cite{PTKP,KRKC,conrfl}. However, the retained data of these methods is continually increased with the number of seen training domains. Besides, these methods also face the risk of data privacy.
Other methods without rehearsal usually constrain its ability to acquire new knowledge to balance the model's stability and plasticity. Thus, the performance of these methods often exhibits suboptimal when confronted with new datasets and is typically inferior to the strategy of data rehearsal.

Different from existing works, in this paper, we propose using textual description as guidance to encourage the ReID model to learn domain-invariant features. The key insight is to construct a shared textual space by describing pedestrian images in an invariant style. This shared space serves as an anchor to guide the ReID model in mapping image features into a unified semantic space.  This could prevent the ReID model from overfitting to specific data domains and alleviate the risk of catastrophic forgetting.
Moreover, mapping all pedestrians into this unified semantic space helps the ReID model mitigate the effects of domain shift, thereby enhancing its generalization capability.

To achieve our objective, we introduce a Dynamic Textual Prompt (DTP) framework, consisting of three key components: the Dynamic Prompt Fusion (DPF) module, the Text-Visual Feature Alignment (TFA) module, and the Learnable Knowledge Distillation (LKD) module.
The DPF module constructs textual prompts to guide the ReID model in embedding pedestrian images into a unified semantic space. 
We generate textual prompts by utilizing a pre-trained captioning module to create pedestrian descriptions and distill existing knowledge from CLIP~\cite{clip-reid} using learnable vectors. These prompts are dynamically integrated using a fusion module.
Besides, in the TFA module, we align the visual features of each local pedestrian region with the textual features of corresponding local phrases sequentially. This fine-grained alignment between visual features and textual prompts ensures effective guidance for embedding all pedestrians into a unified semantic space.
Moreover, the LKD module plays a crucial role by producing adaptive trade-off parameters for learning without forgetting (lwf) loss. This helps strike a balance between learning from new incoming data and preventing catastrophic forgetting.

The experimental results clearly demonstrate that our DTP method achieves superior performance, even outperforming data rehearsal methods by a significant margin. Specifically, our DTP surpasses state-of-the-art competitors by $15.9\%/14.8\%$ over mAP/rank1 on seen datasets and $17.4\%/17.9\%$ over mAP/rank1 on unseen datasets across four training orders. 
The main contributions are summarized as,
\begin{itemize}
\item[$\bullet$] We introduce a novel dynamic textual prompt framework for the LReID task, effectively mitigating catastrophic forgetting by mapping image features with varying distributions into a unified semantic space.
\item[$\bullet$] In our DTP model, we propose a dynamic prompt fusion module to generate textual prompts adaptively. Additionally, a text-visual feature alignment module is introduced to align image features and textual prompts for better fine-grained feature representation.
\item[$\bullet$] Extensive experiments demonstrate the effectiveness of our method, outperforming the state-of-the-art methods on four protocols.
\end{itemize}

\section{Related Works}

\subsection{Person Re-Identification}
Person re-identification (ReID) focuses on matching images of the same pedestrian captured by different cameras~\cite{zheng2017person,PCB,8953719,qian2019leader,qian2018pose}.
Existing ReID methods can be divided into feature representation learning~\cite{zheng2017person,PCB,qian2017multi}, deep metric learning~\cite{wang2019ranked,sun2020circle}, and ranking optimization learning~\cite{8953719,zhang2020understanding}. However, these works often need to be more balanced to the specific datasets and commonly need more capacity for lifelong learning.
Feature representation learning methods~\cite{zheng2017person, PCB,yao2019deep} aim to extract robust and discriminative features from pedestrian images.
Deep metric learning methods~\cite{wang2019ranked,oh2016deep,sun2020circle} aim to design various loss functions to regulate the distance between positive and negative image pairs, aiming to minimize the distance between positive sample pairs while maximizing the distance between negative sample pairs.
During the testing phase of ranking optimization methods, a similarity ranking is introduced~\cite{8953719,sarfraz2018pose}, and the ranking list is then re-optimized based on the similarities between images.
LReID differs from conventional ReID tasks, such as Domain Generalization (DG) or Domain Adaptation (DA). DG~\cite{DG1,DG2,DG3} focuses on acquiring a model that can effectively generalize on unknown datasets, while DA~\cite{DA1,DA2} seeks to minimize the domain gap between source and target domains. In contrast, the LReID task emphasizes dynamic incremental learning across domains and addresses catastrophic forgetting.

\subsection{Lifelong Learning}
Lifelong Learning aims to learn from continuous data streams, so the key challenge is catastrophic forgetting during the learning stages~\cite{dytox,codapromt}.
Lifelong learning methods currently fall into four distinct categories: regularization-based methods, model structure-based methods, data rehearsal-based methods, and prompt-based methods.
Regularization-based methods~\cite{ewc,9157789} focus on constraining the deviation of significant weights within the model during the training phase.
Model structure-based methods~\cite{pnn,bns} facilitate the acquisition of new knowledge by expanding the model's capacity throughout the training process. Data rehearsal-based methods~\cite{KRKC} involve sampling representative samples of old data, which are reused during the current training stage when new data is coming.
Prompt-based methods~\cite{dytox,codapromt} rely on prompt pools comprising numerous learnable prompts, which employ an image as a query, identify the suitable prompt, and utilize it to guide the model for anti-forgetting.

Our work, for the first time, introduces prompts in LReID. Compared with prompt-based methods in classification tasks, our method dynamically fuses two kinds of prompt to instance-wise prompts, which are more focused on fine-grained information than existing class-wise prompts, and our method is more suitable for the LReID task.

\subsection{Prompt-based Architectures}

In natural language processing (NLP), prompt learning has been widely used and developed, with the continuous development of pre-trained models in recent years. The existing model capacity is constantly growing, and the existing NLP paradigm has shifted from ``pre-training and fine-tuning" to ``pre-training, prompting, predicting." By using a given prompt template~\cite{liu2023pre}, the model makes predictions about the outcome in context.
Prompt learning has achieved significant results on several NLP tasks, improving the adaptability of existing pre-trained models to new tasks. CoOP~\cite{coop} introduces prompt learning into computer vision tasks and achieves remarkable results. Prompt learning performs very well in image generation, object detection, and other tasks. 
CLIP-REID~\cite{clip-reid} introduces prompt learning into pedestrian re-identification tasks and uses learnable prompts to guide the model in adapting to pedestrian re-identification tasks.
L2P~\cite{l2p} uses learnable prompts to guide the pre-trained model in adapting to downstream tasks. 
LGCL~\cite{LGCL} maps the features of the continual learner to a semantic space and introduces language guidance to alleviate catastrophic forgetting and obtain a more robust continual learner. 
CODA-prompt~\cite{codapromt} calculates the attention score to multiply the components of prompt pools, changing the previous discrete method to a continuous method.
Our study's critical insight lies in leveraging textual prompts generated by shared large language models (LLMs) to establish an invariant anchor domain. By guiding the ReID model to map image features from diverse domains into this invariant anchor domain, we could prevent the model from overfitting to any particular distribution and thus mitigate the risk of catastrophic forgetting. 
Unlike existing textual and semantic prompting works, our textual prompts are not responsible for incorporating extra knowledge but serve as invariant guidance to force the ReID model to extract domain-invariant features. Our idea is innovative and has the potential to offer valuable insights for a broader range of works in invariant learning.

\subsection{Lifelong Person Re-Identification}
Lifelong Person Re-identification (LReID) ~\cite{AKA,KRKC} seeks to continuously update the model using new data, aiming to sustain the model's performance on previous data distributions while assimilating the knowledge of new distributions.
Each individual can be regarded as a distinct class, emphasizing the significance of discerning alterations in inter-class relationships among different persons.
Moreover, a substantial portion of the pedestrians in the test set have yet to be encountered by the model during its training phase. Consequently, LReID necessitates the capacity to generalize effectively to new categories. 
LReID needs to focus on more fine-grained features to distinguish each individual due to the high similarity among persons and effectively generalize to new categories, especially as the test pedestrians have yet to be encountered during training.
Most existing LReID methods\cite{AKA,KRKC,MEGE,PTKP}  rely on a data rehearsal strategy to preserve historical training data. However, this approach deviates from the initial LReID intent; it demands significant memory resources and may be vulnerable to privacy breaches in practical deployment.
To address this issue, we introduce a novel DTP framework without rehearsal. Our DTP leverages textual prompts to guide the mapping of images to a unified semantic space, thus avoiding the storage of unnecessary and sensitive personal data.

\subsection{Knowledge Distillation}
Knowledge distillation aims to transfer knowledge from the teacher model to the student model by minimizing the gap between their logits and constraining the output distribution of the student model to mimic that of the teacher model. Current knowledge distillation methods typically employ logits-based\cite{jin2023multi,zhao2022decoupled}, feature-based\cite{chen2022knowledge,guo2023class}, and relationship-based approaches\cite{li2022knowledge,yang2021knowledge}.
However, existing LReID methods\cite{AKA,KRKC,MEGE} often utilize a single distillation temperature coefficient across different domains and sharing a global temperature coefficient may not effectively facilitate knowledge transfer. To address this limitation, we propose enhancing the knowledge distillation method in this paper. We introduce learnable parameters to dynamically adjust the distillation temperature, allowing for adaptive changes in temperature to achieve better knowledge transfer across different domains.
\begin{figure*}[h]
  \centering
  \includegraphics[width=1\linewidth]{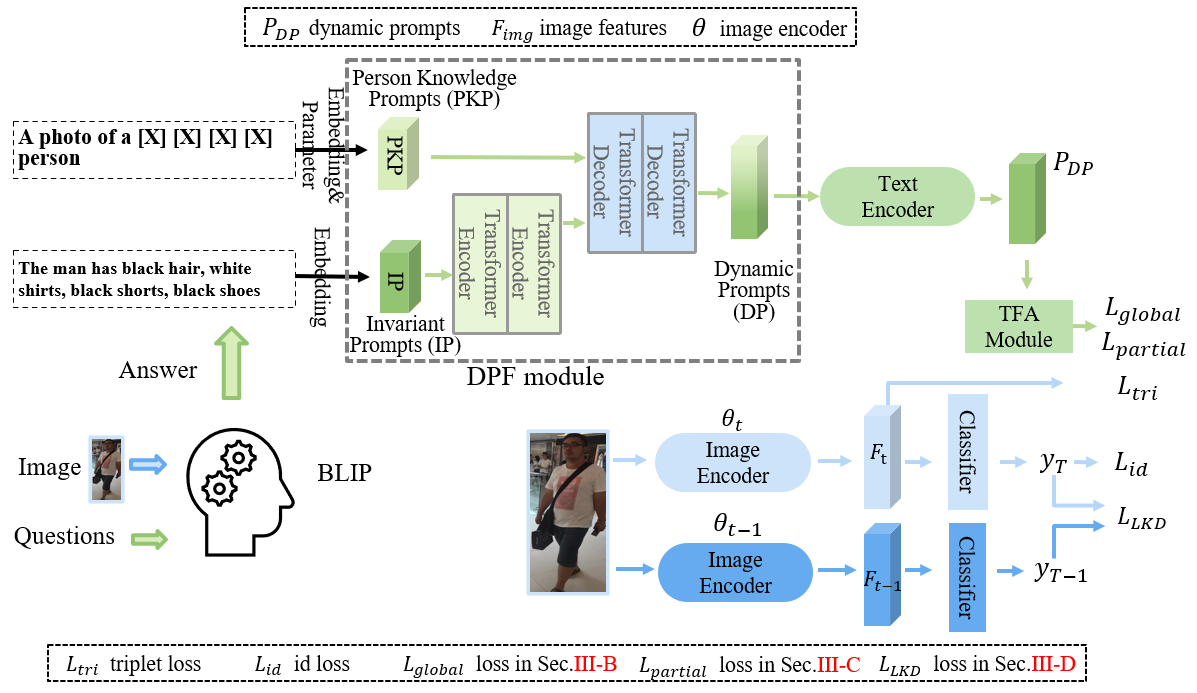}
   \caption{The proposed DTP framework comprises two main modules: Dynamic Prompt Fusion (DPF) and Text-Visual Feature Alignment (TFA). Initially, the DPF module dynamically fuses Invariant Prompts (IP) and Person Knowledge Prompts (PKP) as anchors in semantic space to establish the mapping between the image features and textual prompts. Subsequently, the introduced TFA module aligns local image features and textual prompts between fine-grained local features for a more refined mapping process. Moreover, the LKD module adjusts the temperature coefficient in the knowledge distillation process, dynamically balancing the plasticity and stability of the model.}
   
   \label{fig:framework}
\end{figure*}
\section{Method}
\subsection{Formulation and Method Overview}
\noindent\textbf{Formulation.} The purpose of LReID is to learn an anti-forgetting model from a series of data streams $D={\{D_{train}^{t},D_{test}^{t}\}}^{t=1,...,T}$. 
In stage $T$, the currently accessible dataset is denoted as $D_{T}$, which comprises both the training set $D_{train}^{T}$, and the testing set $D_{test}^{T}$. We train the model on $D_{train}^{T}$ sequentially, and then test it on the test set of all seen datasets, which could be denoted as $\{D_{test}^{1}, ... D_{test}^{T-1}\}$.
After the completion of training in the $T$ stage, the $D_{train}^{T}$ and previous training set are inaccessible. Additionally, it is imperative to note that the training datasets after the $T$ stage remain inaccessible during the $T$ stage. After training, the LReID model needs to maintain good performance on both seen datasets $D={\{D_{test}^{t}\}}_{seen}^{T}$ and unseen datasets $D={\{D_{test}^{t}\}}_{unseen}^{T}$.
\vspace{0.1in}

\noindent \textbf{Method Overview.}
We propose a Dynamic Textual Prompt (DTP) framework for LReID. As depicted in Fig.~\ref{fig:framework}, our DTP comprises three novel components, \textit{i.e.}, Dynamic Prompts Fusion (DPF) module, Text-Visual Feature Alignment (TFA) module and Learnable Knowledge Distillation (LKD) module.
In the DPF module, we adaptively fuse two kinds of prompts to obtain dynamic prompts, guiding the model mapping images of different domains into a unified semantic space. For a more refined and precise mapping, we propose the TFA module to facilitate the alignment of fine-grained textual prompts and local images. 
Besides, in our DTP model, we adopt the learning without forgetting (lwf) loss to avoid catastrophic forgetting.
Here, we design the LKD module to produce a set of adaptive trade-off parameters to balance learning from incoming data and avoiding forgetting.




\subsection{Dynamic Prompt Fusion Module} \label{sec:DPF}
To align image features from different domains into a shared semantic space using various prompts, including natural language descriptions of pedestrians, will guide the image features to a domain-independent text feature space. Integrating multiple prompts to establish anchors for domain-independent mapping poses a significant challenge. To address this challenge, we propose a Dynamic Prompt Fusion (DPF) module. This module combines multiple prompts into dynamic prompts that function as anchors for domain-independent mapping. These dynamic prompts guide images from different domains into a common semantic space and prevent the model from overfitting to the current domain's data distribution.

We utilize BLIP~\cite{BLIP} to pre-process dataset $D_{train}$ to generate the text description of the image and then apply tokenizer and token embedding of CLIP~\cite{CLIP} to extract IP based on these text descriptions. Expressly, we submit the images to BLIP to inquire about the color and style of the clothing.
Then we combined the acquired attributes according to the body parts from top to bottom, such as \emph{``What color is the upper body of the person in the image wearing?"}. Different body parts are divided by the ``,''. Subsequently, we amalgamate the responses to create person descriptions, for example, \emph{``The man has black hair, white T-shirts, black shorts, black shoes."}. In this way, we can get short descriptions for the current datasets. Then, these descriptions are fed into the tokenizer and token embedding of CLIP~\cite{CLIP} to derive IP. 

However, depending solely on IP might overlook subtle discriminative features among closely resembling individuals. To address this, we use the learnable prompt proposed by CLIP-ReID\cite{clip-reid}, which contains a trainable token specific to each ID, providing a more comprehensive and trainable source of textual prompts. We refer to it hereafter as Person Knowledge Prompts (PKP).

Meanwhile, the salience for IP and PKP varies across different instances. While specific samples can attain robust semantic information solely from IP, others—especially individuals with remarkably similar appearances—may exhibit identical textual IP, necessitating additional information. In contrast, PKP tends to generate domain-specific prompts, lacking domain-generalizable knowledge. Introducing IP becomes essential to impart inter-domain generalized knowledge, effectively alleviating the propensity for overfitting to the current task. 
To benefit from IP and PKP simultaneously, we dynamically fuse two kinds of prompts to get Dynamic Prompts (DP).

Specifically, a two-layer transformer encoder extracts features from the IP. Subsequently, both the PKP and IP features are transmitted to the transformer decoder. We leverage the cross-attention mechanism to fuse the two prompts dynamically.
\begin{equation}
  P_{DP}=\Phi(\Psi(P_{IP}),P_{PKP}).
  \label{eq:Prompt_dynamic}
\end{equation}
where $P_{DP}$ represents the dynamic prompts, $P_{IP}$ denotes the invariant prompts, $P_{PKP}$ signifies the person knowledge prompts, $\Psi$ are parameters of transformer encoders, and $\Phi$ indicates parameters of transformer decoders.

We employ $P_{DP}$ as the anchor in semantic space, achieving the mapping of images from diverse domains to a unified semantic space by minimising the cosine similarity between image features and text prompts. Concretely, the $P_{DP}$ is input into the text encoder of CLIP to extract text features. Simultaneously,  the image encoder of CLIP extracts image features. The cosine embedding loss minimises the dissimilarity between the image features and textual prompts. The formulation is articulated as follows:

\begin{equation}
  L_{global}=L_{sup}(P_{DP}\times F_{img}^{T})+L_{sup}(F_{img}\times P_{DP}^{T}).
  \label{eq:kd}
\end{equation}
where $L_{sup}$ is the supconv loss~\cite{supcon}, which is a revised version of cross-entropy loss that uses labels to adjust the distance between positive and negative samples, $F_{img}$ indicates the image feature, $P_{DP}^{T}$ is the transpose of $P_{DP}$, and $F_{img}^{T}$ is the transpose of $F_{img}$.
\begin{figure}
  \centering
  \includegraphics[width=1\linewidth]{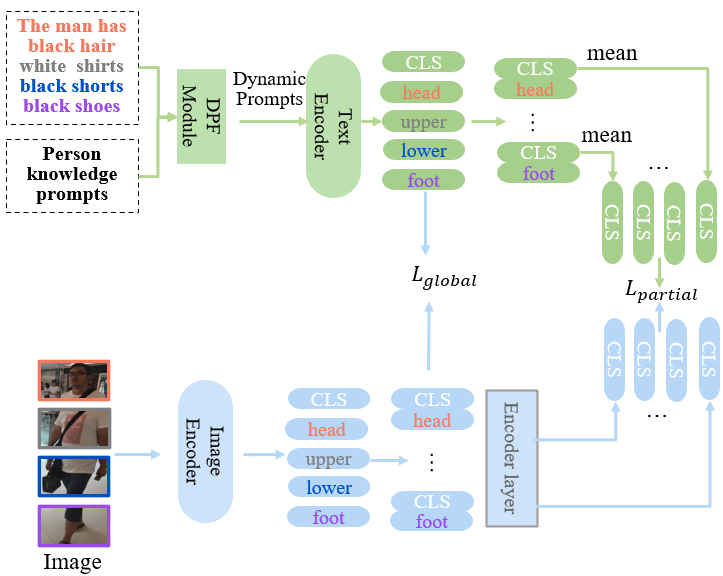}
   \caption{The illustration of the Text-Visual Feature Alignment (TFA) module. The TFA module divides the image features into four blocks according to height and slices the text features according to the different body texts, achieving finer detail alignment by aligning local text and images.}
   \label{fig:tfa}
\end{figure}
\begin{algorithm}
    \caption{Framework of Dynamic Textual Prompt (DTP).}
    \label{alg:algorithm}
    \textbf{Input}: The datasets $D={\{D_{train}^{t},D_{test}^{t}\}}^{T}$ of different domains.\\
    \textbf{Output}: The rank list of $D_{test}$.
    
    \begin{algorithmic}[1] 
        \STATE Let $t=0$.
        \FOR{\textnormal{each datasets in $D$}}
        \STATE Initialize the PKP and parameters of classifier.
        \STATE Get the image captions of $D_{train}^{t}$ from BLIP.
        \STATE \textbf{Stage-I:}
            \FOR{i in range(epochs)}
                \STATE Get DP by DPF module;
                \STATE Get image features $F_{img}$ by fixed image encoder;
                \STATE Get text features $P_{DP}$ by text encoders;
			\STATE Compute the $L_{stage-I}$ and update text encoder;
            \ENDFOR

        \STATE \textbf{Stage-II:}

        \FOR{i in range(epochs) }
            \STATE Get dynamic prompts (DP) by Dynamic Prompt Fusion (DPF) module;
            \STATE Get $F_{img}$ by image encoder;
            \STATE Get $P_{DP}$ by text encoders;
            \STATE Compute the $L_{partial}$ and $L_{global}$ with $P_{DP}$ and $F_{img}$ by TFA module;
            \STATE Compute the $L_{tri}$ and $L_{id}$ with $F_{img}$;
            \STATE Compute the $L_{LKD}$ by $\hat{Y}_{T}$ and $\hat{Y}_{T-1}$;
            \STATE Update the model with $L_{stage-II}$;
        \ENDFOR
        \ENDFOR
        \STATE \textbf{return} the model of resistance to catastrophic forgetting.
    \end{algorithmic}
\end{algorithm}
\subsection{Text-Visual Feature Alignment Module}
In addition to the established global mapping between the image features and textual prompts, we aim for a finer-grained local mapping of image features to textual prompts to achieve more precise alignment mapping. To accomplish this, we design the Text-Visual Feature Alignment (TFA) module that partitions holistic image features and textual prompts into distinct local components and achieves finer-grained alignment. 
This fine-grained alignment could facilitate our model mapping images from diverse domains into a unified semantic space. 

\begin{table*}[h]
\caption{We demonstrate the details for evaluating performance on both seen and unseen domains. For each dataset involved in the experiment, we enumerate the number of identities used for training and testing under balanced and unbalanced protocols.\label{tab:data}}
\centering
\begin{tabular}{c|c|ccc|ccc|c|c|ccc}
\hline
 \multirow{3}*{Domain} & \multirow{3}*{Datasets} & \multicolumn{3}{c|}{Balanced Protocol} & \multicolumn{3}{c|}{Imbalanced Protocol} &\multirow{3}*{Domain} & \multirow{3}*{Datasets} & & \multirow{2}*{Identities} \\
 
 & & \multicolumn{3}{c|}{Identities} &\multicolumn{3}{c|}{Identities} & &\\
 & & Train&Query&Gallery&Train&Query&Gallery & & &Train&Query&Gallery\\
 \hline

\multirow{6}*{Seen}&Market-1501\cite{market1501}&500	&750&751&751&750&751 & \multirow{6}*{Unseen} &SenseReID\cite{sensereid}&-&521&1718\\
	 &CUHK-SYSU\cite{cuhk-sysu}&500&2900&2900&942&2900&2900 & &PRID\cite{prid}&-&100&649\\
	&DukeMTMC\cite{dukemtmc}&500&702&1110&702&702&1110 & &GRID\cite{grid}&-&125&126\\
	&MMT17\cite{msmt17}&500&3060&3060&1041&3060&3060 & &i-LIDS\cite{ilids}&-&60&60\\
	&CUHK03\cite{cuhk03}&500&700&700&767&700&700 & &CUHK01\cite{cuhk01}&-&486&486 \\
	&VIPeR\cite{Viper}&-&-&-&316&316&316 & &CUHK02\cite{cuhk02}&-&239&239 \\
 \hline

\hline
\end{tabular}
\end{table*}
In our TFA module, for image features, inspired by the field of partial re-identification~\cite{PAT,PPCL}, we segment the person image feeatures into four distinct regions from top to bottom, namely, the head, upper body, lower body, and foot features, respectively. 
Subsequently, the hidden features $F_{img}$ before the last layer are evenly partitioned into four segments $f_{part}$ along the sequence length, resulting in $f_{head}$, $f_{upperbody}$, $f_{lowerbody}$, and $f_{foot}$. Each segment is merged with the corresponding cls token and sent into the duplicated encoder layer to extract final local features.
The process can be described as follows,
\begin{equation}
  F_{img}= [f_{head};f_{upper body};f_{lower body};f_{foot}].
  \label{eq:four_part}
\end{equation}

\begin{equation}
  F_{part}=Layer_{part}([cls,f_{part}]).
  \label{eq:four_part_feat}
\end{equation}

The $F_{part}=Layer_{part}$ is the duplicated encoder layer for extracting local features. As for the textual prompts, the generated descriptions are sequentially combined based on body parts, like `The man wears a grey hat, white T-shirt, black shorts, black shoes', so we follow the comma as the segmentation to divide the textual prompts $P_{DP}$ into four parts, \textit{i.e.}, $p_{head}$, $p_{upperbody}$, $p_{lowerbody}$, and $p_{foot}$, which are corresponding to different image semantics. We merge the cls token with each local text feature, averaging by sequence length dimension and generating the final textual local features. We supplement the absent detailed information by PKP in Sec.~\ref{sec:DPF}. This approach ensures a consistent textual pattern that strictly aligns with the four localized images.
The process can be described as follows:

\begin{equation}
  P_{DP}=[p_{head};p_{upper body};p_{lower body};p_{foot}].
  \label{eq:four_part_text}
\end{equation}

With the local textual and visual features, we then enhance fine-grained text-image mapping based on local features by shrinking the cosine distance between image features and textual prompts.
The above process can be formulated as follows:
\begin{equation}
  L_{partial}=\frac{1}{N}\sum^{i}_{N} 1- \frac{p_{i}\cdot f_{i}}{\left\|p_{i}\right\|_2 \cdot \left\| f_{i}\right\|_2}.
  \label{eq:four_part_loss}
\end{equation}
where $f_{i}$ and $p_{i}$ represent the local features of the body part, $L_{partial}$ represents the cosine distance between local image features and textual prompts, $N$ is the number of body parts, and $\left\|*\right\|_2$ represents the L2 norms.
This process enables an effective association of local image features and textual prompts, thereby facilitating model to embed images from diverse domain into an unified semantic space.

\subsection{Learnable Knowledge Distillation}
Knowledge distillation strategy is widely employed to address the catastrophic forgetting problem in LReID. 
Existing works\cite{LwF} suggest that the knowledge distillation strategy is effective in helping the model preserve pre-existing knowledge when learning on the current domain.
However, most of the existing works adopt a fixed trade-off strategy between preserving existing knowledge (stability) and learning new knowledge (plasticity), which is suboptimal.
When facing new and challenging tasks, the model requires more plasticity to learn valuable knowledge, whereas when dealing with simple tasks, more stability is necessary to prevent overfitting.

Therefore, in this work, we propose the learnable knowledge distillation module to balance the plasticity and stability dynamically.
Specifically, The knowledge distillation method\cite{LwF} can be expressed by the formula:

\begin{equation}
L_{KD}=-\sum_{i}y_{T}^{'(i)}\log{y^{'(i)}_{T-1}}.\label{eq:7A}
\end{equation}
\begin{equation}
    y^{'(i)}_{T}=\frac{(y^{(i)}_{T})^{1/t}}{\sum_{j}(y^{(j)}_{T})^{1/t}},\ 
y^{'(i)}_{T-1}=\frac{(y^{(i)}_{T-1})^{1/t}}{\sum_{j}(y^{(i)}_{T-1})^{1/t}}.\label{eq:7B}
\end{equation}
where$ L_{KD}$ denote the knowledge distillation loss, $y_{T}$ denote the identity classification probability vector (after softmax) predicted by the image encoder in the current stage, and $y_{T-1}$ represent the identity classification probability vector of the stored image encoder trained in the previous stage.

To achieve adaptive balancing between preserving pre-existing knowledge and learning new one, we introduce two learnable parameters $\delta1$ and $\delta2$ to control the effort of $L_{KD}$.
The process can be formulated as follows:
\begin{equation}
L_{LKD}=-\sum_{i}\hat{Y}_{T}^{'(i)}\log{\hat{Y}^{'(i)}_{T-1}}.\label{eq:8A}
\end{equation}
\begin{equation}
\hat{Y}^{'(i)}_{T}=\frac{(y^{(i)}_{T})^{1/(t+\delta1)}}{\sum_{j}(y^{(j)}_{T})^{1/(t+\delta1)}},
\hat{Y}^{'(i)}_{T-1}=\frac{(y^{(i)}_{T-1})^{1/(t+\delta2)}}{\sum_{j}(y^{(j)}_{T-1})^{1/(t+\delta2)}}.\label{eq:8B}
\end{equation}

where $L_{LKD}$ represents learnable knowledge distillation loss, $\delta1$ denotes learnable temperature of student model and $\delta2$ indicates learnable temperature of teacher model.

\subsection{Optimization}
As the descriptions in Alg.\ref{alg:algorithm}, the optimization process of the DTP framework is bifurcated into two distinct stages. 
In the first stage, the focus lies in training the text encoder to comprehend the distribution of $P_{DP}$, thereby reducing the risk of an untrained text encoder misleading the image encoder.
This stage targets explicitly the DPF module while freezing the parameters of other modules. The optimization objective is to minimize the distance between the DP and fixed image features, facilitating the establishment of the mapping from image features to textual prompts. The loss function for stage-I $L_{stage-I}$ is $L_{global}$ for optimization.

In the second stage, we train all modules jointly. This optimization objective minimizes the distance between the image features and fixed textual prompts while acquiring pedestrian knowledge in the current domain. The loss function $L_{stage-II}$ for stage-II is described as:
\begin{equation}
  L_{stage-II}=L_{id}+L_{tri}+L_{global}+L_{partial}+0.1L_{LKD}.
  \label{eq:loss_stage2}
\end{equation}
Where $L_{id}$ represents cross entropy loss~\cite{id_loss}, $L_{tri}$ represents triplet loss~\cite{triplet_loss}.

\begin{table*}[h]
  \caption{\textbf{Results for Order-1:} We train the model with the order of MA$\rightarrow$SY$\rightarrow$DU$\rightarrow$MS$\rightarrow$CU03, and test all seen and unseen datasets after the last training stage of CU03 is completed. Additionally, $*$ demonstrates the data-replay methods.\label{tab:p1}}
  \centering
  \begin{tabular}{c|cc|cc|cc|cc|cc|cc|cc}
    \toprule
     \multirow{2}*{Method} & \multicolumn{2}{c|}{MA} & \multicolumn{2}{c|}{SY} & \multicolumn{2}{c|}{DU}& \multicolumn{2}{c|}{MS} & \multicolumn{2}{c|}{CU03} & \multicolumn{2}{c|}{Seen-Avg} & \multicolumn{2}{c}{Unseen-Avg}\\
        & mAP &rank1& mAP &rank1 & mAP &rank1& mAP &rank1& mAP &rank1& $\bar{S}_{mAP}$ & $\bar{S}_{R1}$ & $\bar{S}_{mAP}$ &$\bar{S}_{R1}$\\
    \hline
    CRL (WACV21)~\cite{CRL}&$58.0$&$78.2$&$72.5$&$75.1$&$28.3$&$45.2$&$6.0$&$15.8$&$37.4$&$39.8$&$40.5$ &$50.8$  & $44.0$ &$41.0$\\ 
    AKA (CVPR21) ~\cite{AKA}&$51.2$&$72.0$&$47.5$&$45.1$&$18.7$&$33.1$&$16.4$&$37.6$&$27.7$&$27.6$ &$32.3$ &$43.1$&$44.3$&$40.4$\\
    PTKP\footnotemark[1] (AAAI22)~\cite{PTKP}&$50.3$&$74.8$&$75.4$&$78.0$&$41.2$&$61.5$&$9.8$&$26.3$&$31.7$&$34.1$&$41.7$&$54.9$ & $48.8$ & $44.5$\\
    MEGE (TPAMI23)~\cite{MEGE} &$39.0$ &$61.6$&$73.3$  &$76.6$&$16.9$&$30.3$&$4.6$&$13.4$&$36.4$&$37.1$&$34.0$&$43.8$ &$47.7$ &$44.0$ \\
    ConRFL(PR23)~\cite{conrfl}& $59.2$ & $78.3$ & $82.1$ & $84.3$ &$45.6$&$61.8$&$12.6$&$30.4$&$51.7$&$53.8$& $50.2$ & $61.7$ & $57.4$ & $52.3$\\
    LSTKC(AAAI24) & 54.7 &76.0 &81.1 & 83.4 & 49.4 & 66.2 & 20.0 &43.2 &44.7 &46.5 &50.0 &63.1 & 57.0 & 49.9 \\
    
    DTP (Ours)
    & $\textbf{69.1}$ & $\textbf{85.4}$ & $\textbf{90.5}$ & $\textbf{91.6}$ & $\textbf{65.4}$ & $\textbf{79.6}$ & $\textbf{39.5}$ & $\textbf{66.4}$ & $\textbf{73.3}$ & $\textbf{75.5}$ & $\textbf{67.6}$ & $\textbf{79.7}$ & $\textbf{77.9}$ & $\textbf{72.1}$
    \\
    \bottomrule
  \end{tabular}
\end{table*}
\begin{table*}
\caption{\textbf{Results for Order-2:} We train the model with the order of DU$\rightarrow$MS$\rightarrow$MA$\rightarrow$SY$\rightarrow$CU03, and test all seen and unseen datasets after the last training stage of CU03 is completed. Additionally, $*$ demonstrates the data-replay methods.\label{tab:p2}}
  \centering
  \begin{tabular}{c|cc|cc|cc|cc|cc|cc|cc}
    \toprule
     \multirow{2}*{Method} & \multicolumn{2}{c|}{DU}& \multicolumn{2}{c|}{MS} & \multicolumn{2}{c|}{MA} & \multicolumn{2}{c|}{SY}  & \multicolumn{2}{c|}{CU03} & \multicolumn{2}{c|}{Seen-Avg} & \multicolumn{2}{c}{Unseen-Avg}\\
        & mAP &rank1& mAP &rank1 & mAP &rank1& mAP &rank1& mAP &rank1& $\bar{S}_{mAP}$ &$\bar{S}_{R1}$ & $\bar{S}_{mAP}$ &$\bar{S}_{R1}$\\
    \hline 
    CRL (WACV21)~\cite{CRL} & $43.5$ & $63.1$ & $4.8$ & $13.7$ & $35.0$ & $59.8$ & $70.0$ & $72.8$ & $34.8$ & 36.8 & $37.6$ & $49.2$  &$43.9$ &$40.1$\\
    AKA (CVPR21) ~\cite{AKA} & $42.2$ & $60.1$ & $5.4$ &$ 15.1$ & $37.2$ & $59.8$ & $71.2 $&$ 73.9$ &$ 36.9$ & $37.9$ & $38.6$ & $49.4$ & $40.6$ &$34.2$\\
    PKD(ACM MM22) ~\cite{PKD} & $58.3$ &$ 74.1$ & $6.4$ & $17.4$ & $43.2$ & $67.4$ & $74.5$ &$ 76.9$ & $33.7$ & $34.8$ & $43.2$ & $54.1$ &$48.6$ &$44.1$ \\
    PTKP\footnotemark[1] (AAAI22)~\cite{PTKP} & $34.2$&$52.2$&$7.3$&$19.8$&$55.0$&$78.8$&$79.9$&$81.9$&$30.8$&$41.6$&$41.4$&$54.9$&$48.6$ &$43.9$\\
    MEGE (TPAMI23) ~\cite{MEGE} &$21.6$&$35.5$ &$3.0$&$9.3$&$25.0$&$49.8$&$69.9$&$73.1$&$34.7$&$35.1$&$30.8 $&$40.6$ &$44.3$ &$41.1$\\
    ConRFL\footnotemark[1](PR23)~\cite{conrfl} &$34.4$&$51.3$&$7.6$&$20.1$&$61.6$&$80.4$&$82.8$&$85.1$&$49.0$&$50.1$&$47.1 $&$57.4$ &$57.9$ &$53.4$ \\
    LSTKC(AAAI24) & 49.9&67.6&14.6&34.0&55.1&76.7&82.3&83.8&46.3&48.1&49.6&62.1&57.6&49.6\\
    
    DTP (Ours)
    & $\textbf{60.2}$ &$ \textbf{75.0}$ & $\textbf{31.4}$ & $\textbf{57.9}$ & $\textbf{78.4}$ & $\textbf{90.0}$ & $\textbf{92.3}$ & $\textbf{93.2}$ & $\textbf{74.1}$ & $\textbf{76.7}$ & $\textbf{67.3}$ & $\textbf{78.6}$ & $\textbf{76.7}$ & $\textbf{70.1}$
    \\
    \bottomrule
  \end{tabular}
\end{table*}

\section{Experiments}

\subsection{Experiments settings}
\noindent{\textbf{Datasets.}} Two groups of datasets are utilized in our experiments: LReID model seen and LReID model unseen datasets. \textbf{1) LReID seen datasets:} Market1501 (MA)~\cite{market1501}, CUHK-SYSU (SY)~\cite{cuhk-sysu}, MSMT17 (MS)~\cite{msmt17}, CUHK03 (CU03)~\cite{cuhk03}, DukeMTMC (DU)~\cite{dukemtmc} and VIPeR (VIP)~\cite{Viper}. \noindent\textbf{2) LReID unseen datasets:} VIPeR~\cite{Viper}, prid~\cite{prid}, grid~\cite{grid}, ilids~\cite{ilids}, CUHK01~\cite{cuhk01}, CUHK02~\cite{cuhk02}, sensereid~\cite{sensereid}. Noted, Whether the VIPeR dataset belongs to seen or unseen is determined by the benchmarks.

\noindent{\textbf{Evaluation metrics.}}
Mean Average Precision (mAP) and Rank-1 accuracy (rank1) are computed to evaluate each task. Additionally, the average incremental accuracy, denoted as $\bar{S}_{R1}$, and $\bar{S}_{mAP}$, are assessed using the mean average precision and Rank-1 accuracy as measures.~\cite{AKA,conrfl,MEGE}

\noindent{\textbf{Training orders.}}
For a fair comparison, we follow the training orders of existing LReID works~\cite{MEGE,conrfl,KRKC} and report the performance of seen and unseen datasets after completing the training stage over the last seen dataset.

\textbf{1) Balanced orders:} Following prior work, we restrict the usage of 500 unique IDs within the datasets, commonly referred to as balanced orders~\cite{AKA,conrfl,MEGE}. 
\textbf{\textit{Order-1:}} The training order is MA$\rightarrow$SY$\rightarrow$DU$\rightarrow$MS$\rightarrow$CU03. The unseen test datasets are VIPeR, prid, grid, ilids, CUHK01, CUHK02, and sensereid.
\textbf{\textit{Order-2:}} The training order is DU$\rightarrow$MS$\rightarrow$MA$\rightarrow$SY$\rightarrow$CU03. The unseen test datasets are similar to Order-1.
\textbf{\textit{Order-3:}} The training order is MA$\rightarrow$SY$\rightarrow$MS$\rightarrow$CU03. The unseen test datasets are the same as Order-1.

{\textbf{2) Imbalanced orders:} Following previous work~\cite{PTKP,gwfreid,KRKC}, we use all person ID for training.
\textbf{\textit{Order-4:}} The training orders is VIP$\rightarrow$MA$\rightarrow$SY$\rightarrow$MS. The unseen test datasets are CUHK01, CUHK02, CUHK03, and prid.
After training is completed, we directly test the model on all seen and unseen datasets and report in Tab.\ref{tab:p1}-Tab.\ref{tab:p4}.

\noindent{\textbf{Implementation Details.}}
All person images are standardized to the dimensions of $256\times128$. Each image undergoes augmentation through random horizontal flipping, padding, cropping, and erasing.
For optimization, we employ the Adam optimizer with a learning rate of $3.5\times10^{-4}$, initiating a warm-up learning rate of $0.01$, utilizing a linear warm-up approach. The training epochs of stage-I and stage-II are both $60$. A weight decay $1e$-$4$ is applied, and the batch size is $128$, with loss weights set to 1 during the training stage.
The model architecture is founded on the CLIP~\cite{CLIP}, comprising a text and image encoder. To adjust CLIP's original position encoding from $[197, 768]$ to $[129, 768]$, bilinear interpolation is employed. 

\begin{table*}
    \caption{\textbf{Results for Order-3:} We train the model with the order of MA$\rightarrow$SY$\rightarrow$MS$\rightarrow$CU03, and test all seen and unseen datasets after the last training stage of CU03 is completed. Additionally, $*$ demonstrates the data-replay methods.\label{tab:p3}}
  \centering
  \setlength{\tabcolsep}{8pt}
  \begin{tabular}{c|cc|cc|cc|cc|cc|cc}
    \toprule
     \multirow{2}*{Method} & \multicolumn{2}{c|}{MA} & \multicolumn{2}{c|}{SY} & \multicolumn{2}{c|}{MS}& \multicolumn{2}{c|}{CU03}& \multicolumn{2}{c|}{Seen-Avg}& \multicolumn{2}{c}{Uneen-Avg}\\
         & mAP &R1& mAP &rank1 & mAP &rank1& mAP &rank1& $\bar{S}_{mAP}$ &$\bar{S}_{R1}$ & $\bar{S}_{mAP}$ &$\bar{S}_{R1}$\\
    \hline 
     AKA(CVPR21)~\cite{AKA} & 44.4 & 67.7 & 73.8 & 79.8 & 4.1 & 11.8 & 33.9 & 34.0 & 39.1 & 48.3 & 46.5 & 43.1 \\
    MRN(ACM MM22)~\cite{MRN} & 57.6 & 80.6 & 77.5 & 79.8 & 16.5 & 39.9 & 42.9 & 43.7 & 48.6 & 61.0 & 56.1 & 52.7 \\
    Ours  & 
    $\textbf{69.4}$ & $\textbf{84.8}$ & $\textbf{90.9}$ & $\textbf{92.0}$ & $\textbf{38.6}$ & $\textbf{64.6}$ & $\textbf{73.1}$ & $\textbf{74.9}$ & $\textbf{68.0}$ & $\textbf{79.1}$ & $\textbf{76.2}$ & $\textbf{70.1}$ 
    \\
    \bottomrule
  \end{tabular}

\end{table*}

\begin{table*}
    \caption{\textbf{Results for Order-4:} We train the model with the order of VIP$\rightarrow$MA$\rightarrow$SY$\rightarrow$MS, and test all seen and unseen datasets after the last training stage of MS is completed. Additionally, $*$ demonstrates the data-replay methods.\label{tab:p4}}
  \centering
  \setlength{\tabcolsep}{8pt}
  \begin{tabular}{c|cc|cc|cc|cc|cc|cc}
    \toprule
     \multirow{2}*{Method} & \multicolumn{2}{c|}{VIP} & \multicolumn{2}{c|}{MA} & \multicolumn{2}{c|}{SY}& \multicolumn{2}{c|}{MS}& \multicolumn{2}{c|}{Seen-Avg}& \multicolumn{2}{c}{Unseen-Avg}\\
         & mAP &rank1& mAP &rank1 & mAP &rank1& mAP &rank1& $\bar{S}_{mAP}$ &$\bar{S}_{R1}$ & $\bar{S}_{mAP}$ &$\bar{S}_{R1}$\\
    \hline
    AKA (CVPR21)\cite{AKA}&$61.7$&$50.6$&$28.3$&$50.7$&$76.9$&$79.6$&$13.4$&$28.0$&$45.1$&$52.2$ &$32.6$&-\\
    PTKP \footnotemark[1](AAAI22)\cite{PTKP}&$66.3$&$56.0$&$58.3$&$77.4$&$77.1$&$78.8$&$25.2$&$48.0$&$56.7$&$65.1$ &$42.2$ &-\\ 
    KRKC \footnotemark[1](AAAI23)\cite{KRKC}&$76.3$&$67.7$&$64.4$&$82.5$&$88.9$&$90.7$&$43.3$&$67.1$&$68.2$&$77.0$&$60.3$&-\\

    DTP (Ours) 
     & $\textbf{79.6}$&$\textbf{70.9}$&$\textbf{ 70.3}$&$\textbf{87.3}$&$\textbf{92.9}$&$\textbf{94.1}$ &$\textbf{66.4}$ & $\textbf{84.0}$ & $\textbf{77.3}$ & $\textbf{85.3}$ &$\textbf{70.7}$ &$\textbf{68.8}$ 
    \\
    \bottomrule
  \end{tabular}
\end{table*}

\subsection{Comparisons with SOTAs}
\noindent{\textbf{We can achieve the best results on seen domains.}}
We demonstrate the effectiveness of our DTP method in Tab.~\ref{tab:p1}-Tab.~\ref{tab:p4}. Our DTP significantly outperforms all competitors, which leads to a $14.8\%$ increase in $\bar{S}_{R1}$ and a $15.9\%$ increase in $\bar{S}_{mAP}$ on four seen datasets from four training orders.
Our DTP can retain previously acquired knowledge and improve learning by mapping images with diverse domains into a unified semantic space. Thanks to serving the textual prompts as anchors, DTP can promote the learning of anti-forgetting image features and effectively solve the forgetting problem. 

\noindent{\textbf{We can achieve the best results on unseen domains.}}
In addition to performing well on seen datasets, our DTP performs well on all unseen data sets in Tab.~\ref{tab:p1}-Tab.~\ref{tab:p4}. The results demonstrate the efficacy of DTP on unseen settings, and the DTP module can boost the performance of $17.9\%$ over $\bar{S}_{R1}$ and $17.4\%$ over $\bar{S}_{mAP}$ on different unseen orders.
The experimental results unequivocally illustrate our model's efficacy under various unseen protocols. It adeptly maps distinct image features from diverse unseen domains into a unified semantic space for learning. Consequently, DTP demonstrates a robust generalization capability, even when applied to datasets it has not encountered before, underscoring its notable advantages on unseen domains.

\subsection{Ablation study}

\begin{table*}
\caption{Ablation studies on each component of our DTP. The Baseline is the CLIP-ReID. The ‘Addition’ represents the element-wise addition of IP and PKP. For CLIP-ReID, we replaced the prompt on each visible dataset during training.\label{tab:effectiveness}}
  \centering
  \setlength{\tabcolsep}{9pt}
  \begin{tabular}{c|cccccc|cccc}
    \toprule
     \multirow{2}*{ID} & \multirow{2}*{Baseline~\cite{clip-reid}}& \multirow{2}*{Addition}& \multirow{2}*{DPF} & \multirow{2}*{TFA} & \multirow{2}*{KD} & \multirow{2}*{LKD}
     & \multicolumn{2}{c}{Seen} & \multicolumn{2}{c}{Unseen} \\
         & & & & & & &\multicolumn{1}{c}{$\bar{S}_{mAP}$}& $\bar{S}_{R1}$ &\multicolumn{1}{c}{$\bar{S}_{mAP}$} & $\bar{S}_{R1}$ \\
    \hline
    1&\Checkmark & & & & & & $63.9$ & $76.1$ &$74.1$ &$66.7$ 
    \\
    2&\Checkmark &\Checkmark & & & & & $ 64.2 $ & $ 76.3 $ & $ 74.9 $ & $ 68.2 $  
    \\
    3&\Checkmark & &\Checkmark & & & & $65.0$ & $76.7$ & $76.4$ & $70.0$ \\
    
    4&\Checkmark & &\Checkmark &\Checkmark & & & $66.6$ & $77.9$ & $76.6$ & $70.0$ \\
    5&\Checkmark & &\Checkmark &\Checkmark &\Checkmark & &$67.0$ & $78.1$ &$76.7$ &$69.9$ \\
    6&\Checkmark & &\Checkmark &\Checkmark &\Checkmark &\Checkmark & $\textbf{67.3}$ & $\textbf{78.6}$ & $\textbf{76.7}$ & $\textbf{70.1}$\\
    \bottomrule
  \end{tabular}
\end{table*}

In this section, we conduct extensive experiments to demonstrate the effectiveness of DTP from various aspects, \textit{i.e.}, the effectiveness of critical components, the impact of balanced and imbalanced datasets, the forgetting tendency, different hypermeters, and visualization results.

\noindent{\textbf{The effectiveness of each component in DTP framework.}}
We conduct sufficient experiments to examine each DTP component's effectiveness in Tab.~\ref{tab:effectiveness}. All the experiments are evaluated in Order-2.
From the results of Tab.~\ref{tab:effectiveness}, we can notice the following observations, 

\noindent\textit{(1) Evaluation of our DPF module.}
(a) We conduct additional experiments to assess the impact of two distinct prompts. 
Comparing the ID1 and ID2, upon introducing IP, we observe the performance increase of $0.9\%$/$0.6\%$ over $\bar{S}_{R1}$/$\bar{S}_{mAP}$, demonstrating that simply adding IP and PKP does not lead to an exemplary mapping of image features to textual semantics.
(b) We conduct an investigation into the effects of dynamic fusion, which led to a consistent improvement in the $1.1\%$/$1.2\%$ on $\bar{S}_{R1}$/$\bar{S}_{mAP}$ metrics by changing the element-wise addition to dynamic prompt fusion strategy as shown in ID2 and ID3. This finding underscores the ability of instance-wise prompts acquired through dynamic fusion to guide each sample toward a more accurate adaptive prompt;
(c) The results of ID1 and ID4 demonstrate the impact of the DPF module, which yields the enhancement in the $2.0\%$/$1.7\%$ over $\bar{S}_{R1}$/$\bar{S}_{mAP}$ compared to the image encoder.

\noindent\textit{(2) Investigation of the TFA module.}
As shown in ID4 and ID5, the introduction of the TFA module resulted in an improvement of $1.2\%$/$1.6\%$ over $\bar{S}_{R1}$/$\bar{S}_{mAP}$ metrics of seen datasets. Our findings suggest that the knowledge transfer facilitated by fine-grained local features can effectively guide a more refined process of image-text mapping.

\noindent\textit{(3) Exploration of the LKD module.}
As shown in ID5 and ID6, the introduction of the LKD module resulted in an improvement of $0.5\%$/$0.3\%$ over $\bar{S}_{R1}$/$\bar{S}_{mAP}$ on seen datasets. During the distillation of knowledge, the knowledge of the past can be better remembered through the dynamic adjustment of the distillation temperature.

\noindent\textit{(4) Comparison between the CLIP-ReID and our DTP.} 
Notably, our method's performance improvement is attributed to the robustness of the CLIP-ReID approach and the effectiveness of our proposed DTP frameworks. 
The DPF module dynamically fuses the two prompts to map the image features to semantic space, while the TFA module achieves fine-grained mapping between local features; the LKD module adjusts the plasticity and stability of the model during knowledge distillation.
As shown in ID1 and ID6, our DTP can achieve enhancements over CLIP-ReID, yielding a $3.4\%$ increase in $\bar{S}_{mAP}$ and a corresponding improvement of $2.5\%$ of $\bar{S}_{R1}$ on seen and a $2.6\%$ increase in $\bar{S}_{mAP}$ and a corresponding improvement of $3.3\%$ of $\bar{S}_{R1}$ unseen setting.

\begin{table}
\caption{Experiments were conducted on balanced and unbalanced domains. Our method exhibited superior performance.
\label{tab:imbalance_unseen}}
\begin{center}
\setlength{\tabcolsep}{4pt}
 \renewcommand{\arraystretch}{1.0}
  \begin{tabular}{c|c|cccc|cc}
    \hline
    \multirow{2}*{Order}& \multirow{2}*{Protocol} & \multicolumn{2}{c}{AKA} & \multicolumn{2}{c|}{MEGE} & \multicolumn{2}{c}{Ours}\\
         & & $\bar{S}_{mAP}$& $\bar{S}_{R1}$ &$\bar{S}_{mAP}$ & $\bar{S}_{R1}$
         &$\bar{S}_{mAP}$ & $\bar{S}_{R1}$ \\
    \hline
    \multirow{2}*{Order-1}&Balanced & $46.6$ & $43.1$ & $47.7$ & $44.0$ & $\textbf{67.6}$ & $\textbf{79.7}$\\
    &Imbalanced & $54.0$ & $50.5$ & $55.1$ &$51.3$ & $ \textbf{70.4}$ & $\textbf{82.2}$\\ 
    \hline
    
    \multirow{2}*{Order-2}&Balanced& $43.7$ & $40.8$ & $44.3$ &$41.1$ & $\textbf{67.3}$ & $\textbf{78.6}$\\
    &Imbalanced & $51.2$ & $48.2$ &$53.2$ &$50.4$ & $\textbf{68.1}$ & $\textbf{79.1}$   \\ 
    \hline
    
\end{tabular}
\end{center}
\end{table}

\noindent{\textbf{The qualitative experiments on balanced and imbalanced datasets.}}
In this section, over order-1 and order-2 datasets, we report the impact of whether the training datasets are balanced on the performance of our method. Our method still performs best compared with others, as shown in Tab.~\ref{tab:p4} and Tab.~\ref{tab:imbalance_unseen}. Such results demonstrate that our method can map the image features of imbalanced datasets to a common semantic space without data rehearsal, and can continually learn new knowledge even under imbalanced datasets, which is closer to real-world applications. 

\begin{table}[!t]
\caption{Experiments on the performance effects of different prompt fusion methods, simple element-wise addition and concate methods are detrimental to performance on Order-2.\label{tab:fuse}}
\centering
\begin{tabular}{c|cc|cc}
\hline
\multirow{2}*{Method} & \multicolumn{2}{c|}{Seen-Avg}& \multicolumn{2}{c}{Uneen-Avg}\\
         &$\bar{S}_{mAP}$ &$\bar{S}_{R1}$ & $\bar{S}_{mAP}$ &$\bar{S}_{R1}$\\
\hline
Addition& $ 64.2 $ & $ 76.3 $ & $ 74.9 $ & $ 68.2$\\ 
Concate& $61.2$ & $74.1$ & $72.2$ &$66.5$\\
DPF (Ours) &$\textbf{67.3}$ & $\textbf{78.6}$ & $\textbf{76.7}$ & $\textbf{70.1}$\\
\hline
\end{tabular}
\end{table}
\begin{table*}
\caption{Experiments of our methods on more different orders and comparisons with State-of-the-Arts. After the training order is disrupted, our method still achieves the best performance on balanced and unbalanced datasets\label{tab:diff_orders}}
\begin{center}
\setlength{\tabcolsep}{6pt}
\renewcommand{\arraystretch}{1.1}
  \begin{tabular}{c|c|ccc|ccc|ccc|ccc}
    \hline
    \multirow{3}*{Protocol}& \multirow{3}*{Order} &  \multicolumn{6}{c|}{Seen domains} & \multicolumn{6}{c}{Unseen domains}\\
    &  & \multicolumn{3}{c}{$\bar{S}_{mAP}$} &  \multicolumn{3}{c|}{$\bar{S}_{R1}$} & \multicolumn{3}{c}{$\bar{S}_{mAP}$} &  \multicolumn{3}{c}{$\bar{S}_{R1}$} \\
    \cline{3-14}
    &  & AKA & MEGE & Ours  & AKA & MEGE & Ours  & AKA & MEGE & Ours  & AKA & MEGE & Ours \\
    \hline
    \multirow{6}*{Balanced}& MS$\rightarrow$SY$\rightarrow$DU$\rightarrow$MA$\rightarrow$CU03& $33.6$ & $36.2$ & $\textbf{66.8}$  & $43.7$ & $48.3$ & $\textbf{78.2}$  & $48.3$ & $50.1$ & $\textbf{75.5}$ & $44.8$ & $45.9$  & $\textbf{68.7}$ \\ 
        &DU$\rightarrow$MA$\rightarrow$CU03$\rightarrow$MS$\rightarrow$SY & $34.7$ & $38.8$ & \textbf{67.9}  & $45.8$ & $49.7$ & \textbf{78.5} & $40.6$ & $41.9$ & \textbf{75.5} & $37.6$ & $39.0$  & \textbf{69.2} \\
        &SY$\rightarrow$DU$\rightarrow$CU03$\rightarrow$MS$\rightarrow$MA& $37.0$ &$ 42.3$ & \textbf{67.8}  & $45.1$ & $49.4 $& \textbf{77.9} & $38.6$ & $40.2$ & \textbf{75.8} & $35.7$ & $36.3$  & \textbf{69.7} \\
        &CU03$\rightarrow$MS$\rightarrow$DU$\rightarrow$MA$\rightarrow$SY & 34.7 & 38.7 & $\textbf{64.9}$  & 44.6 & 49.1 & $\textbf{74.7}$ & 43.5 & 44.0 & \textbf{75.4} & 39.7 & 40.8  & \textbf{68.8} \\
        &MA$\rightarrow$MS$\rightarrow$DU$\rightarrow$SY$\rightarrow$CU03 & 31.1 & 33.8 & $\textbf{65.9}$  & 41.5 & 43.7 &$\textbf{77.5}$ & 45.6 & 47.3 & $\textbf{76.0}$  & 43.0 & 44.2  & $\textbf{69.4}$ \\
        \cline{2-14}
        & Average & 34.2 & 38.0 & $\textbf{66.7}$  & 44.1 & 48.0 & $\textbf{76.8}$  & 43.3 & 44.7 & $\textbf{75.6}$  & 40.2 & 41.2  & $\textbf{69.2}$ \\
        \hline
         \multirow{7}*{Imbalanced}& MS$\rightarrow$SY$\rightarrow$DU$\rightarrow$MA$\rightarrow$CU03 & 40.8 & 44.6 & $\textbf{70.1}$ & 44.9 & 48.2 & $\textbf{80.6}$ & 54.1 & 55.6 & $\textbf{78.3}$ & 51.2 & 52.3 & $\textbf{72.3}$ \\
        & DU$\rightarrow$MA$\rightarrow$CU03$\rightarrow$MS$\rightarrow$SY & 41.7 & 45.0 & $\textbf{69.2}$ & 45.1 & 49.0 & $\textbf{78.8}$ & 50.2 & 51.3 & $\textbf{78.4}$ & 47.3 & 48.8  & $\textbf{72.5}$ \\
        & SY$\rightarrow$DU$\rightarrow$CU03$\rightarrow$MS$\rightarrow$MA & 46.4 & 49.3 & $\textbf{71.7 }$ & 49.8 & 51.8 & $\textbf{80.5 }$ & 45.7 & 46.5 & $\textbf{77.4}$ & 46.5 & 47.6  & $\textbf{71.0}$ \\
        & CU03$\rightarrow$MS$\rightarrow$DU$\rightarrow$MA$\rightarrow$SY & 45.1 & 48.9 & $\textbf{66.5}$ & 47.1 & 51.3 & $\textbf{75.7 }$  & 52.9 & 54.1 & $\textbf{77.0}$ & 50.5 & 51.8  & $\textbf{71.2}$ \\
        & MA$\rightarrow$MS$\rightarrow$DU$\rightarrow$SY$\rightarrow$CU03 & 40.1 & 43.2 & $\textbf{69.0}$ & 48.5 & 51.9 & $\textbf{80.6}$ & 51.0 & 51.7 & $\textbf{76.7}$ & 49.0 & 49.2  & $\textbf{70.4}$ \\
         \cline{2-14}
        & Average & $42.8$ & $46.2$ & $\textbf{69.3}$ & $47.1$ & $50.4 $& $\textbf{79.2}$ & $50.8$ & $51.8$ & $\textbf{77.6}$ & 48.9 & 49.9  & $\textbf{71.5}$\\
    \hline
\end{tabular}
\end{center}
\end{table*}
{\noindent {\textbf{Exploration over different fusion method.}}
In Tab.\ref{tab:fuse}, we compare our DPF module with other prompt fusion methods, namely, the version element-wise adds PKP and IP together (``Addition''), and the version concatenates PKP and IP (``Concate'') in sequence dimension. To ensure the sequence length of ``Concate'' satisfies the input length of CLIP's text encoder, we apply a pooling operation over its sequence dimension.
Our DPF demonstrates superior performance compared to other methods, achieving a $1.1\%$ improvement in $\bar{S}_{mAP}$ and a $1.2\%$ enhancement in $\bar{S}_{R1}$.

\noindent{\textbf{Investigation on different training orders.}}
To assess the impact of various training orders on our method, we adopted the protocol established by MEGE~\cite{MEGE} and conducted additional ablation experiments across different domain sequences in Tab.~\ref{tab:diff_orders}. The outcomes demonstrate a significant performance enhancement compared to previous methodologies. Specifically, our approach surpasses $25.9\%$ $\bar{S}_{mAP}$ and $28.8\%$ $\bar{S}_{R1}$ on the seen domains and  $28.3\%$ $\bar{S}_{mAP}$ and $24.8\%$ $\bar{S}_{R1}$ on the unseen domains. This indicates a robustness to training order variations, showcasing notable results across diverse domain sequences. Thus, our method effectively addresses the challenge of domain variance.
\begin{figure}[t]
  \centering
  \begin{tabular}{cc}
  \includegraphics[width=0.24\textwidth]{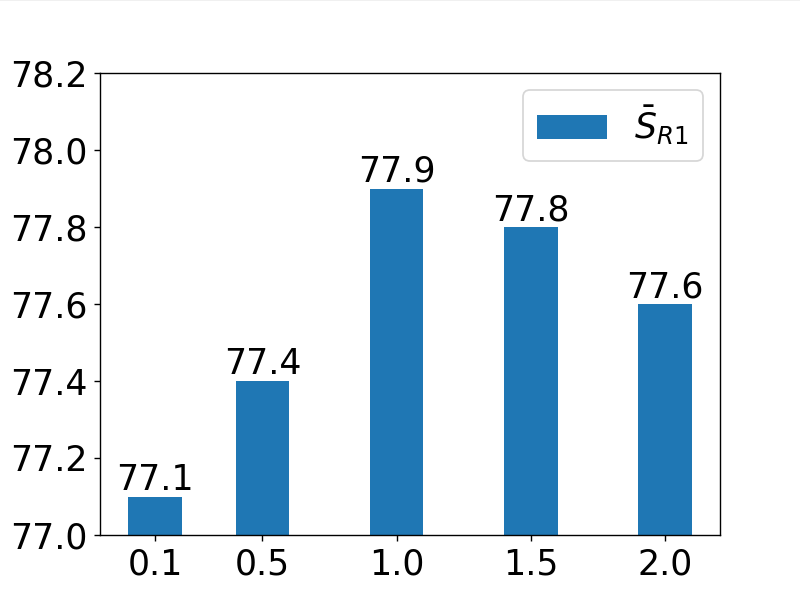}&
     \includegraphics[width=0.24\textwidth]{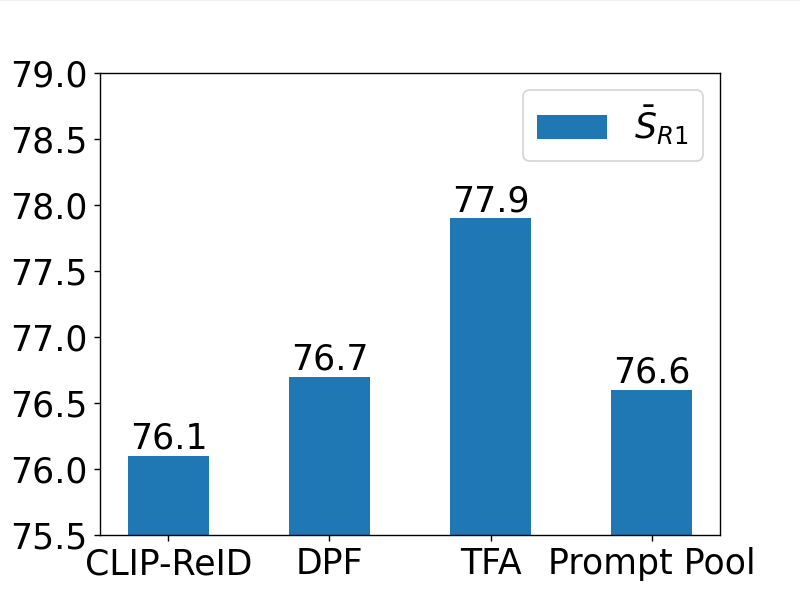}\\
    (a) $\lambda_{TFA}$&(b) Different prompt methods\\
    \includegraphics[width=0.24\textwidth]{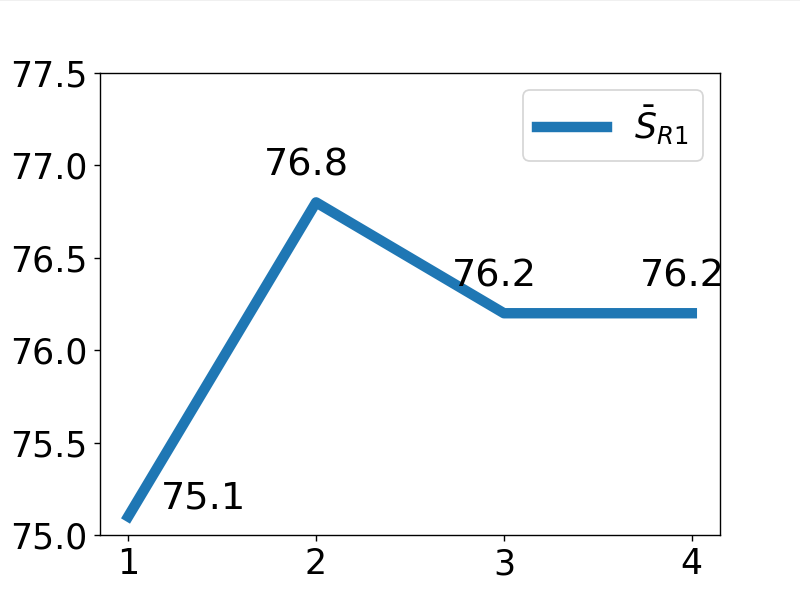}&
     \includegraphics[width=0.24\textwidth]{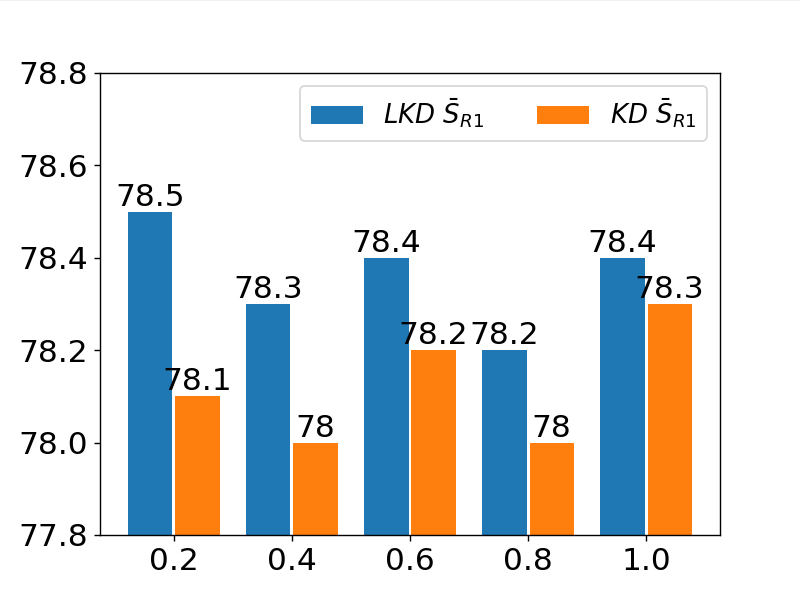}\\
    (c) The number of layers&(d) Different temperature\\
    \end{tabular}
   \caption{The effectiveness of different hypermeters.}
   \label{fig:parameters}
\end{figure}
\begin{figure}
\begin{center}
\begin{tabular}{cc}
\includegraphics[width=0.24\textwidth]{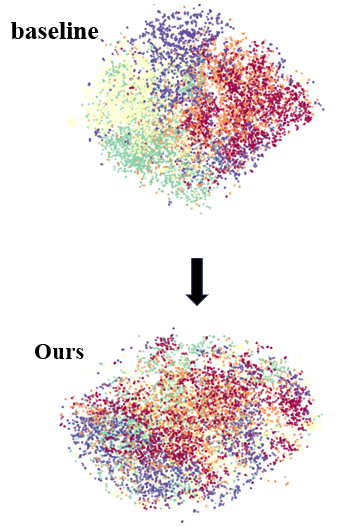}&
\includegraphics[width=0.24\textwidth]{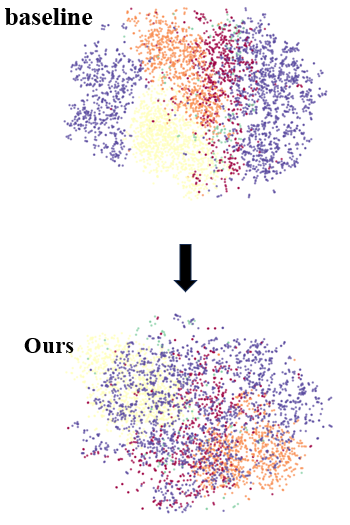}\\
(a)&(b)\\
\end{tabular}
\caption{t-SNE visualization of feature distribution of Order-1, various colours demonstrate different domains. (a) Visualization of the image encoder and ours on seen datasets, the domain gaps between feature distributions are eliminated after applying our method. (b) Visualization of the image encoder and ours on unseen datasets and the results show that our method can even eliminate the domain gaps among unseen domains.
\label{fig:t-SNE}}
\end{center}
\end{figure}

\noindent{\textbf{More experiments over hypermeters.}}
We present the impact of various parameters on Order-2 in Fig.~\ref{fig:parameters}. 

\noindent\textit{(1) The weights of $L_{TFA}$ loss.} The performance of our TFA module is influenced by the weight attributed to $L_{TFA}$ loss. We systematically vary the weight from 0.1 to 2, and its corresponding effectiveness is depicted in Fig.~\ref{fig:parameters}(a). Consequently, we set the $\lambda_{TFA}$ value for all experimental configurations.

\noindent\textit{(2) Different prompt methods.}
We compare different prompt methods in the figure, and the prompt pool method maintains a sizeable prompt pool to memorize knowledge. As shown in Fig.~\ref{fig:parameters}(b), our method performs better in the LREID task.

\noindent\textit{(3) Different numbers of transformer layer.}
We adjust the number from $1$ to $4$, as shown in Fig.~\ref{fig:parameters}(c). 
Increasing the number of transformer layers does not necessarily enhance the model's effectiveness. This suggests that augmenting the training parameters may incline the prompts towards overfitting the current domain distribution, exacerbating the forgetting of prior knowledge. Merely amplifying the grounding parameters of the DPF module fails to enhance the model's anti-forgetting capability. Conversely, when the number of transformer layers within the DPF module is insufficient, optimal integration of the two prompts becomes challenging, thus diminishing the model's capacity to assimilate new knowledge.

\noindent\textit{(4) Different temperature.} In Fig.~\ref{fig:parameters}(d), we further explore the effect of LKD loss at different temperatures. We set the distillation temperature to different values and compared the effect of LKD loss and KD loss at the same base distillation temperature; we can see that our LKD loss is better than the traditional knowledge distillation loss.

\begin{figure*}[htbp]
  \centering
  \begin{tabular}{cccc}
    \includegraphics[width=0.23\textwidth]{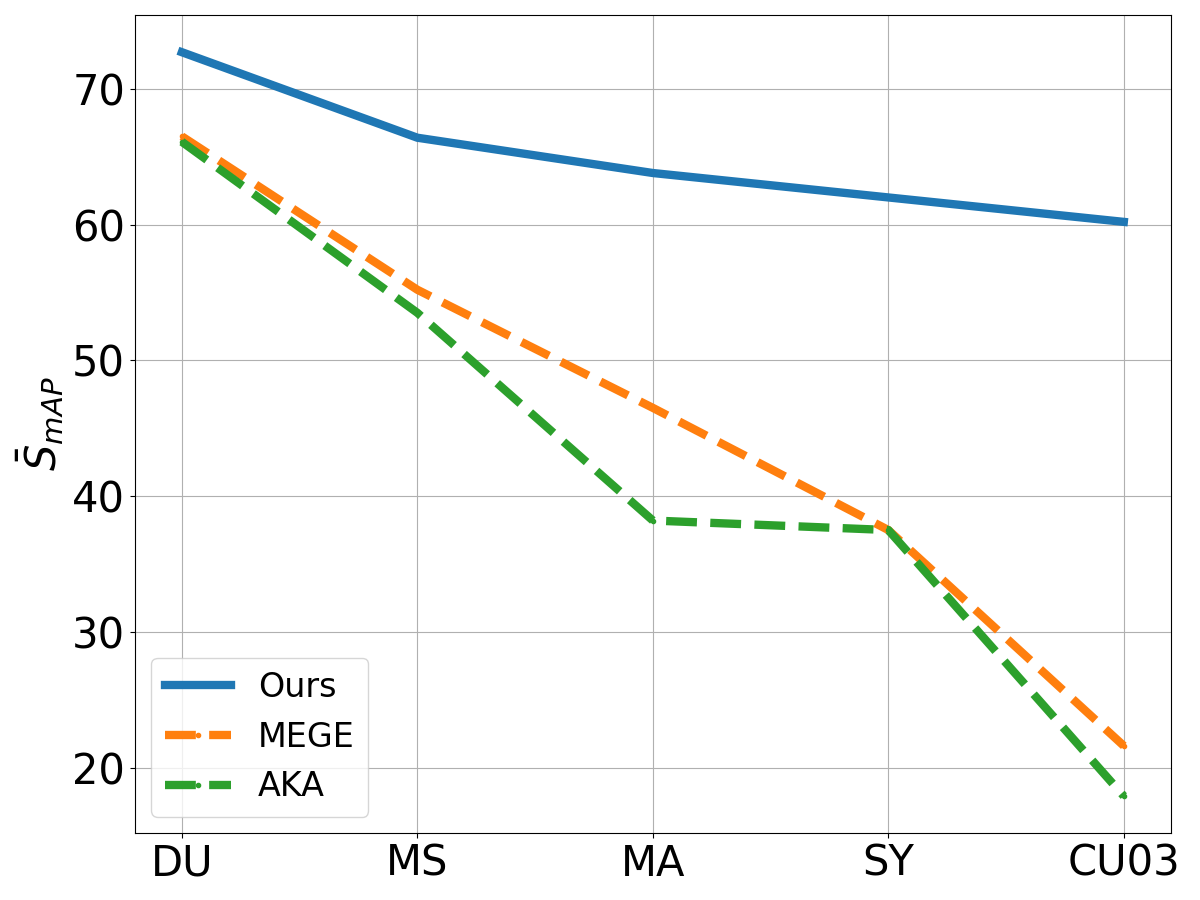}&
    \includegraphics[width=0.23\textwidth]{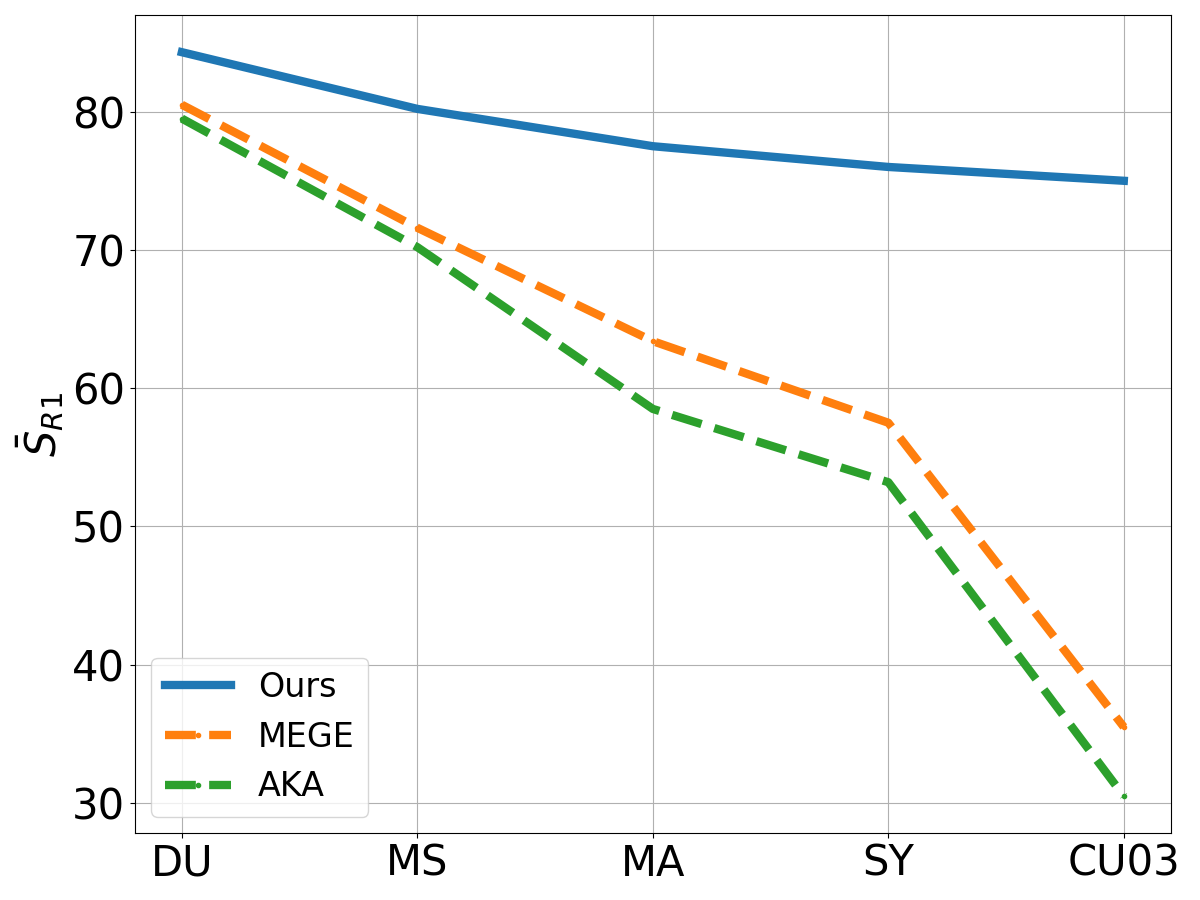}&
    \includegraphics[width=0.23\textwidth]{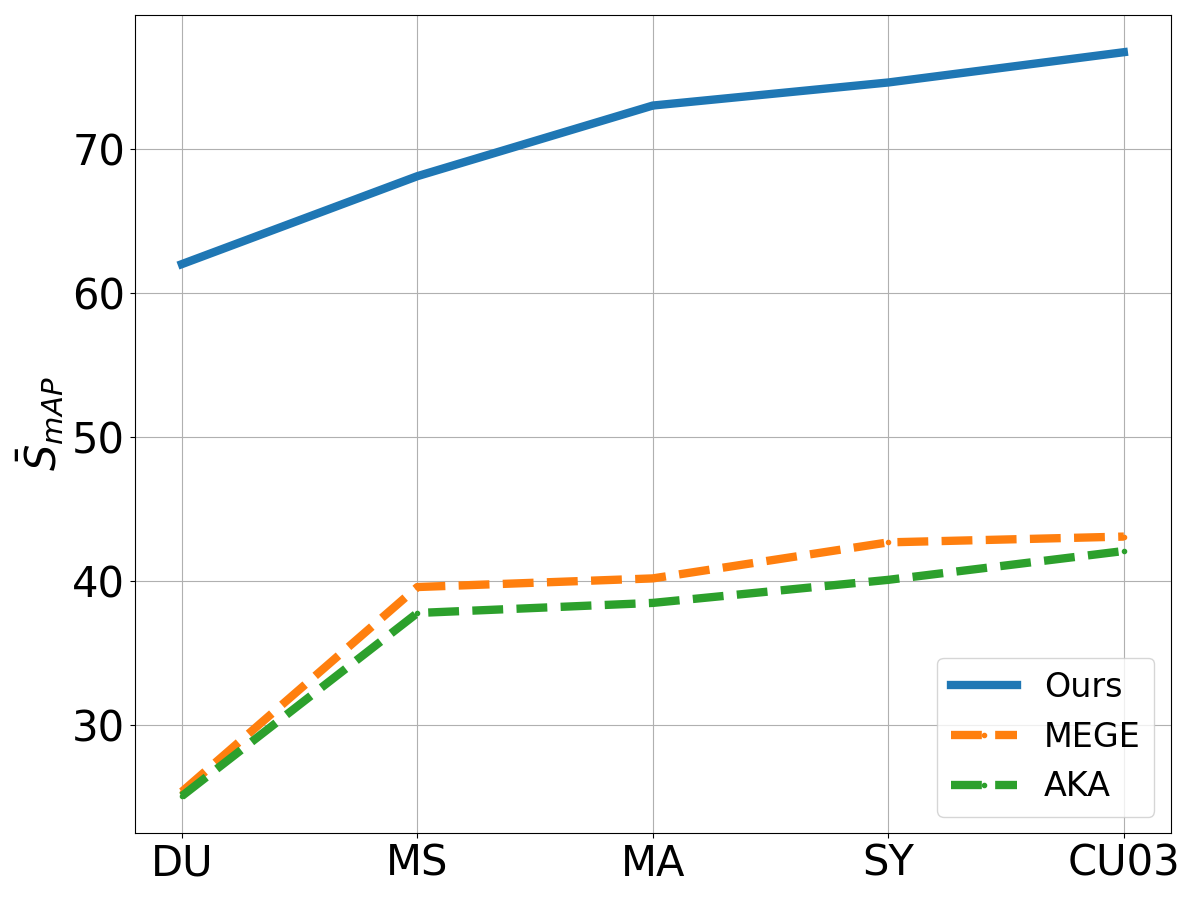}&
    \includegraphics[width=0.23\textwidth]{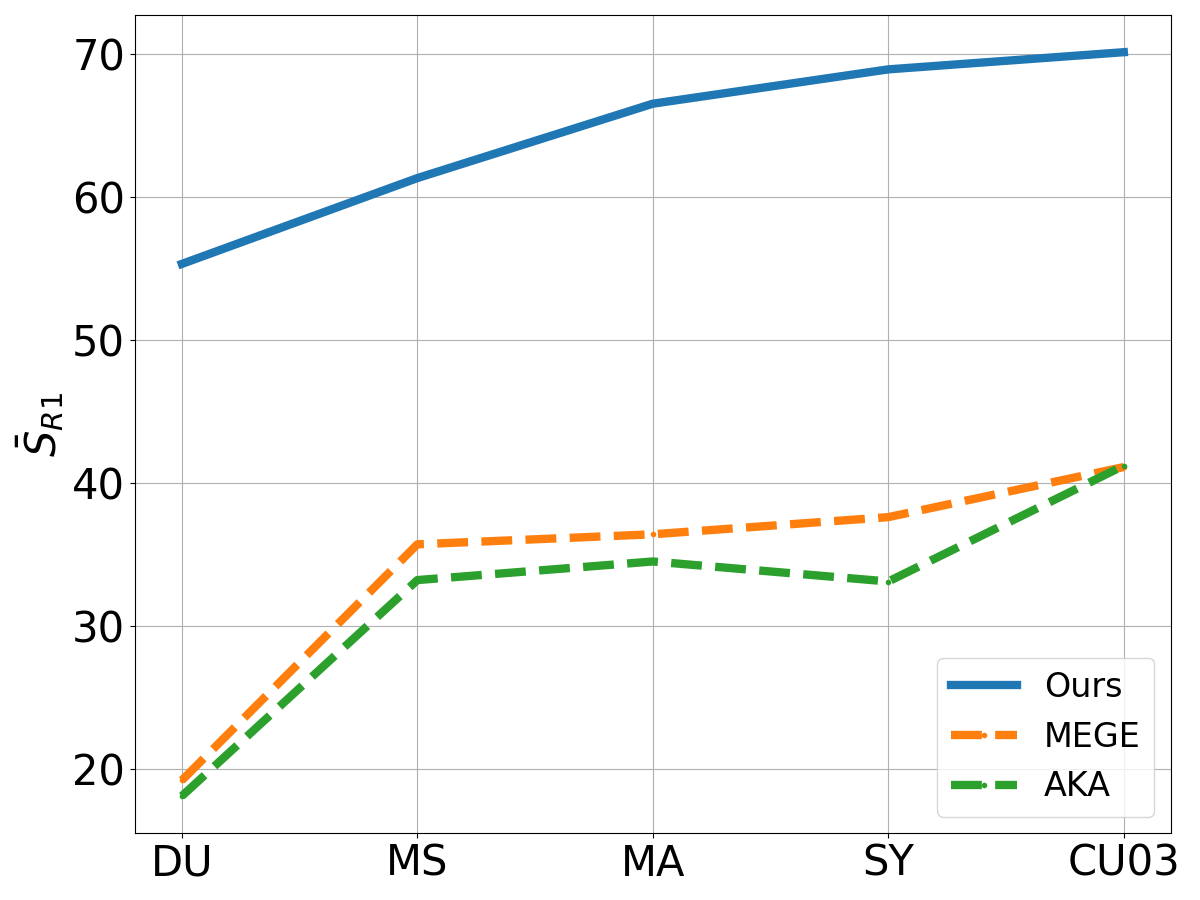}
    \\
    (a) The mAP of Order-2&(b) The rank1 of Order-2 &(c) The mAP of Order-2&(d) The rank1 of Order-2\\
    
    \end{tabular}
    \caption{The comparisons of forgetting tendency on balanced datasets, our method has a smoother forgetting curve compared to previous methods, especially in the later stages of training, and does not suffer from the sharp degradation in performance of previous methods.}
   \label{fig:Forgetting_Tendency}
\end{figure*}

\subsection{Visualization}

\noindent{\textbf{T-SNE results.}}
To illustrate our DTP framework's effectiveness, we compare the t-SNE results for the baseline  (Image Encoder in Tab.~\ref{tab:effectiveness}) and DTP models under Order-1. As shown in Fig.~\ref{fig:t-SNE}, the domain gaps of different distributions of image encoder exhibit distinct boundaries, while our method breaks the boundaries of different feature distributions and eliminates the domain gaps. Notably, our method can even break the boundaries of unseen feature distributions with better generalization capabilities. Such observations indicate that our approach maps image features from diverse domains into a unified semantic space. Consequently, it mitigates the model's susceptibility to overfitting new data distributions during continuous learning, addressing the challenge of catastrophic forgetting.

\noindent{\textbf{Forgetting Tendency.}}
We track the performance of the first seen training dataset of Order-2 after each training stage in Fig.~\ref{fig:Forgetting_Tendency}.
As illustrated in Fig.~\ref{fig:Forgetting_Tendency} (a) and (b), our method exhibits a smoother forgetting curve, significantly distinct from the rapid decline observed in the forgetting curves of MEGE~\cite{MEGE} and AKA~\cite{AKA}. This distinction emphasizes the effectiveness of our approach in mitigating forgetting tendencies.

\noindent{\textbf{Generalizing Tendency:}} We track the average performance of the unseen dataset on Order-2, and report the $\bar{S}_{R1}$ and $\bar{S}_{mAP}$ after each training stage in Fig.\ref{fig:Forgetting_Tendency} (c) and (d). The results of our DTP demonstrate a notable improvement in generalizing tendency over previous methods, showing robust performance across unseen domains. This enhancement can be attributed to our approach to effectively learning standard semantic features across diverse domains.

\section{Conclusion}
This paper introduces the dynamic textual prompt framework for lifelong person re-identification. Our framework guides the model in learning to map images of different domains to a unified semantic space. We propose that the dynamic prompts fusing module be used to dynamically fuse two types of prompts, providing more discriminative dynamic prompts. Furthermore, we introduce a text-visual feature alignment module to align the similarity of fine-grained local features and establish a fine-grained mapping. We also propose the learnable knowledge distillation module to dynamically balance the plasticity and stability.
Ultimately, our methods set a new SOTA on several LReID settings and significantly outperform previous approaches. The extensive experiments demonstrate the efficacy of our method.

\section{Ethical considerations}
As for the ethical considerations of the DukeMTMC dataset, we chose it due to the widely established benchmarks~\cite{AKA,conrfl,MEGE,gwfreid} and fair comparison with existing methods, we ensure it has been used strictly only for research purposes and handled in compliance with the community's standards.
\clearpage

\bibliographystyle{IEEEtran}
\bibliography{TIP24}

\clearpage



\newpage

\vfill

\end{document}